\documentclass[final]{cvpr}

\usepackage{times}
\usepackage{epsfig}
\usepackage{graphicx}
\usepackage{amsmath}
\usepackage{amssymb}
\usepackage{multirow}
\usepackage{tabularx}
\usepackage{xcolor}
\usepackage{lineno}
\usepackage[export]{adjustbox}
\usepackage{appendix}

\newcommand\blfootnote[1]{%
  \begingroup
  \renewcommand\thefootnote{}\footnote{#1}%
  \addtocounter{footnote}{-1}%
  \endgroup
}
\newcommand{\minisection}[1]{\vspace{0.04in} \noindent {\bf #1}}
\newcolumntype{Y}{>{\centering\arraybackslash}X}

\usepackage[pagebackref=true,breaklinks=true,colorlinks,bookmarks=false]{hyperref}

\def\cvprPaperID{25} 
\def\confYear{CLIC 2021}

\begin{document}

\title{DANICE: Domain adaptation without forgetting in neural image compression}

\author{Sudeep Katakol\textsuperscript{1}\footnotemark[1] \thinspace\quad Luis Herranz\textsuperscript{2} \thinspace\quad Fei Yang\textsuperscript{2} \thinspace\quad Marta Mrak\textsuperscript{3} \\[0.4em]  \textsuperscript{1}Univ. of Michigan, Ann Arbor \quad  \textsuperscript{2}Computer Vision Center, UAB, Barcelona \quad
\textsuperscript{3}BBC R\&D, London
}

\maketitle

\begin{abstract}
   Neural image compression (NIC) is a new coding paradigm where coding capabilities are captured by deep models learned from data. This data-driven nature enables new potential  functionalities. In this paper, we study the adaptability of codecs to custom domains of interest. We show that NIC codecs are transferable and that they can be adapted with relatively few target domain images. However, naive adaptation interferes with the solution optimized for the original source domain, resulting in forgetting the original coding capabilities in that domain, and may even break the compatibility with previously encoded bitstreams. Addressing these problems, we propose Codec Adaptation without Forgetting (CAwF), a framework that can avoid these problems by adding a small amount of custom parameters, where the source codec remains embedded and unchanged during the adaptation process. Experiments demonstrate its effectiveness and provide useful insights on the characteristics of catastrophic interference in NIC.
\end{abstract}

\section{Introduction}
\blfootnote{* This work was done during an internship at the CVC, Barcelona.}

Lossy image and video coding have been the cornerstone of visual content sharing and communication, achieving high compression rates by allowing a small amount of distortion. With the recent advances in machine learning with neural networks, neural image compression (NIC)~\cite{toderici2015variable,balle2016end,theis2017lossy,toderici2017full} has emerged as a new powerful paradigm where codecs
optimize their parameters directly using data from the domain of interest, leading to very competitive coding performance. The main limitations for practical deployment are high memory and computation requirements and lack of flexibility. Recent works have addressed some of these practical limitations, including memory and computational complexity~\cite{yang2021slimmable,johnston2019computationally}, and  variable rate~\cite{toderici2015variable, choi2019variable, yang2020variable, cai2018efficient,theis2017lossy}. 
But the NIC paradigm can also enable novel functionalities.

\begin{figure}[t]
	\centering
	\includegraphics[width=\columnwidth]{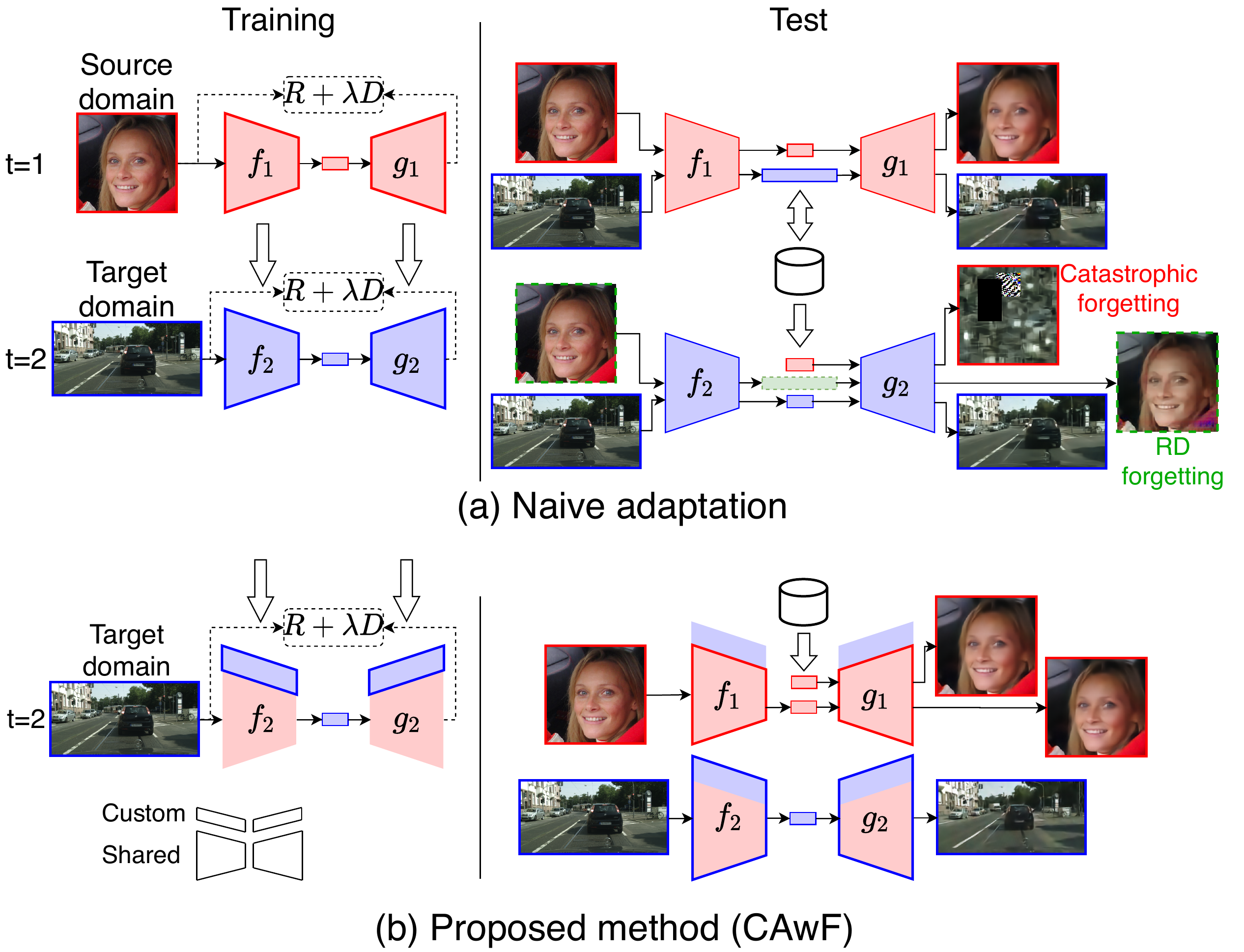}
	
	\caption{Domain adaptation and forgetting in NIC: (a) naive adaptation results in forgetting, while (b) CAwF prevents forgetting.}
	\label{fig:overview}
	\vspace{-0.5em}
\end{figure}

Traditional coding formats (such as those from JPEG~\cite{wallace1992jpeg} and MPEG families~\cite{wiegand2003overview,sullivan2012overview,vvc2020}) follow a carefully designed syntax and reference decoder, which are strictly defined by standards to enable interoperability and backward compatibility. In contrast, NIC codecs discover an implicit latent syntax through learnable parametric models. 
These parameters could be readjusted to improve the performance in a custom domain or functionality. Fig.~\ref{fig:overview} shows possible scenarios involving such type of customization, and illustrates potential compatibility issues. 
For instance, an off-the-shelf learned codec optimized for a certain domain of interest (shown in red at $t=1$ in Fig.~\ref{fig:overview}a) may not necessarily be optimal for another domain, with performance suffering due to higher bitrate and/or higher distortion.
The trainable nature of learned codecs is a convenient mechanism to customize the off-the-shelf codec, resulting in an adapted codec with lower rate and distortion in that domain of interest (shown in blue in Fig. \ref{fig:overview}a at $t=2$).
There are many potential use cases of this functionality, such as privately adapted codecs for photographers' personal collections, or an on-board camera with a codec adapted to the specific vehicle. In traditional codecs, adapting to specific applications would require designing specific tools or even new coding formats. In NIC, this functionality can be reinterpreted as domain adaptation~\cite{wang2018deep}, which in machine learning refers to the reuse and adaptation of the knowledge~\cite{azizpour2015factors} learned in a source model to improve performance in a target domain where data is scarce.

The main limitation of such codec adaptation process is that the new parameters are not optimal for the source domain anymore (phenomenon known as forgetting~\cite{mccloskey1989catastrophic} in continual learning~\cite{parisi2019continual}, see Fig.~\ref{fig:overview}a). This can manifest as lower rate-distortion (RD) performance when encoding images of the source domain (i.e. RD forgetting). This drop in performance is more significant in small models, often required in resource-limited scenarios or when efficiency is a concern. Furthermore,  the codec becomes incompatible with bitstreams encoded with the original version of the codec. While the decoder is able to generate an image, the result is disastrous (i.e. catastrophic forgetting). Thus, preventing forgetting is crucial when the codec must remain compatible and with good performance in the source domain.

In this work, we first study the adaptability of learned codecs. For cases where forgetting is a concern, we then propose an incremental architecture (see Fig.~\ref{fig:overview}b) that prevents both RD and catastrophic forgetting by fixing the source codec and using a small number of additional custom parameters to learn target-specific patterns. In this way, the previous version of the codec remains embedded in the codec and can be used independently when necessary.

In summary, the main contributions of this work are:
\begin{itemize}
\item We introduce the problem of domain adaptation in neural image compression (DANICE), and the related problem of forgetting the original coding capabilities. In this way we connect concepts of traditional image coding and machine learning, such as backward compatibility, domain adaptation and forgetting.

\item We study the adaptability of NIC codecs and propose selective fine tuning for cases with few target images.

\item We characterize how forgetting manifests in NIC codecs, proposing a framework that prevents forgetting by design, while ensuring optimal rate-distortion performance and backward compatibility.
\end{itemize}

\section{Adapting NIC codecs}
\subsection{NIC framework}
	The most common framework consists of a learnable feature autoencoder\footnote{Our autoencoder combines convolutional and (I)GDN~\cite{balle2016density} layers.}, combined with quantization and learnable entropy coding~\cite{balle2016end,theis2017lossy,balle2018variational}. The encoder $\mathbf{b}=f\left(\mathbf{x};\theta,\nu\right)$ maps the input image $\mathbf{x}\in \mathcal{X} \subset \mathbb{R}^{N}$ to the bitstream $\mathbf{b}$. The decoder reconstruct the image as $\mathbf{\hat{x}}=g\left(\mathbf{b};\phi,\nu\right)$. The full model is determined by its parameters $\psi=\left(\theta,\phi,\nu\right)$ (i.e. encoder, decoder, entropy model). 
	
	The parameters $\psi$ are learned by minimizing\footnote{During training, quantization is replaced by a differentiable proxy to allow end-to-end training via backpropagation~\cite{balle2016end}.} a combination of rate and distortion over a training set $\mathcal{X}^\text{tr}$ sampled from the domain of interest $\mathcal{X}$ 
	\begin{gather}
		J\left(\mathcal{X}^\text{tr},\psi;\lambda\right)=R\left(\mathcal{X}^\text{tr},\psi\right)+\lambda D\left(\mathcal{X}^{tr},\psi\right) \label{eq:RD_loss},
	\end{gather}
	where $\lambda$ is the (fixed) tradeoff between rate\footnote{Approximated by the entropy during training.} $R\left(\mathcal{X}^\text{tr},\psi\right)$ and distortion\footnote{Here measured as the average mean square error (MSE).} $D\left(\mathcal{X}^\text{tr},\psi\right)$. See Appendix A for more details.
	
\subsection{Adapting to new domains}

We introduce the problem of \textit{domain adaptation in neural image compression} (DANICE), where a codec trained on a source domain $\mathcal{X}_1$ is leveraged to improve compression in a target domain $\mathcal{X}_2$.

A straightforward approach to DANICE is \textit{naive fine tuning}, where we minimize $J\left(\mathcal{X}_2^\text{tr},\psi;\lambda\right)$ with a target training set $\mathcal{X}^\text{tr}_2$ in order to obtain a target model $\psi_2$. Naive fine tuning can easily lead to overfitting when the number of images is small relative the total number of parameters.

We can mitigate overfitting and improve training stability by reducing the number of tunable parameters. Concretely, we tune the GDN layers~\cite{balle2016density} and the entropy model, while keeping the convolutional layers fixed. We introduce a small number of channel-specific parameters to adjust scales and biases ($\tilde{z}_i=\alpha_i z_i+\beta_i$, for the $i$-th channel of a feature $\mathbf{z}$). Through this \textit{selective fine tuning} approach, parameters that are tuned are reduced to only $\sim 5\%$ of the original amount, improving adaptation under few images.

\section{Forgetting and compatibility}
Now, consider the case where we are interested in adapting a source codec (optimized for $\mathcal{X}_1$) to $\mathcal{X}_2$, whilst retaining good compression performance on images from domain $\mathcal{X}_1$. 
Moreover, we may also want to decode any previously encoded $\mathcal{X}_1$ images, i.e. have backward compatibility.

\subsection{Characterizing forgetting}
This new setting is more challenging than adaptation, because optimizing for the new domain \textit{interferes}~\cite{mccloskey1989catastrophic} with the solution for the source domain, leading to \textit{forgetting}, i.e. the performance of the model in the source domain drops. 

We can distinguish between two cases, leading to two different types of forgetting (described below). For simplicity, we use $f_t$ to refer to the encoder at time $t$, i.e. $f\left(\mathbf{x};\theta_t,\nu_t\right)$. Similarly, we use $g_t$ to refer to $g\left(\mathbf{b};\phi_t,\nu_t\right)$. 

\minisection{Rate-distortion forgetting}
Consider that a source image $\mathbf{x}\in\mathcal{X}_1$ is encoded with the encoder $f_2$ and decoder $g_2$ optimized for the target domain $\mathcal{X}_2$, i.e. $\mathbf{\hat{x}}=g_2\left(f_2\left(\mathbf{x}\right)\right)$. In this case, encoder and decoder belong to $t=2$, so they remain compatible. However, as the codec is optimized for a different domain, forgetting occurs, resulting in new artifacts and a worse rate-distortion performance compared to using $f_1$ and $g_1$  (see Fig.~\ref{fig:overview}a).

\minisection{Catastrophic forgetting}
In this case, forgetting is closely related to backward compatibility.
When an image $\mathbf{x}\in\mathcal{X}_1$ is encoded with $f_1$ and saved, and then decoded with an updated version of the decoder $g_2$, i.e. $\mathbf{\hat{x}}=g_2\left(f_1\left(\mathbf{x}\right)\right)$, the image cannot be recovered at all (see Fig.~\ref{fig:overview}a). 

\subsection{Preventing forgetting and incompatibility}
In order to address the aforementioned problems, we propose \textit{Codec Adaptation without Forgetting} (CAwF), an image coding architecture composed of \textit{shared} and \textit{custom} parameters. The source codec, optimized for the source domain, is stored in the shared parameters, $\psi_1$. The custom parameters, $\Delta\psi$ are used together with shared ones to obtain the target codec, i.e.  $\psi_2=\psi_1\bigcup\Delta\psi$. The adaptation process also minimizes $J\left(\mathcal{X}_2^\text{tr},\psi;\lambda\right)$, but the shared parameters remain fixed. In our case, the custom parameters are distributed across the different layers of the feature autoencoder, and a separate entropy model is learned. This architecture prevents both RD forgetting and interference between domains by design. The user can choose the version of the encoder, which is always signalled in the bitstream.

\begin{figure}
\vspace{-0.8em}
    \centering
    \includegraphics[width=\columnwidth]{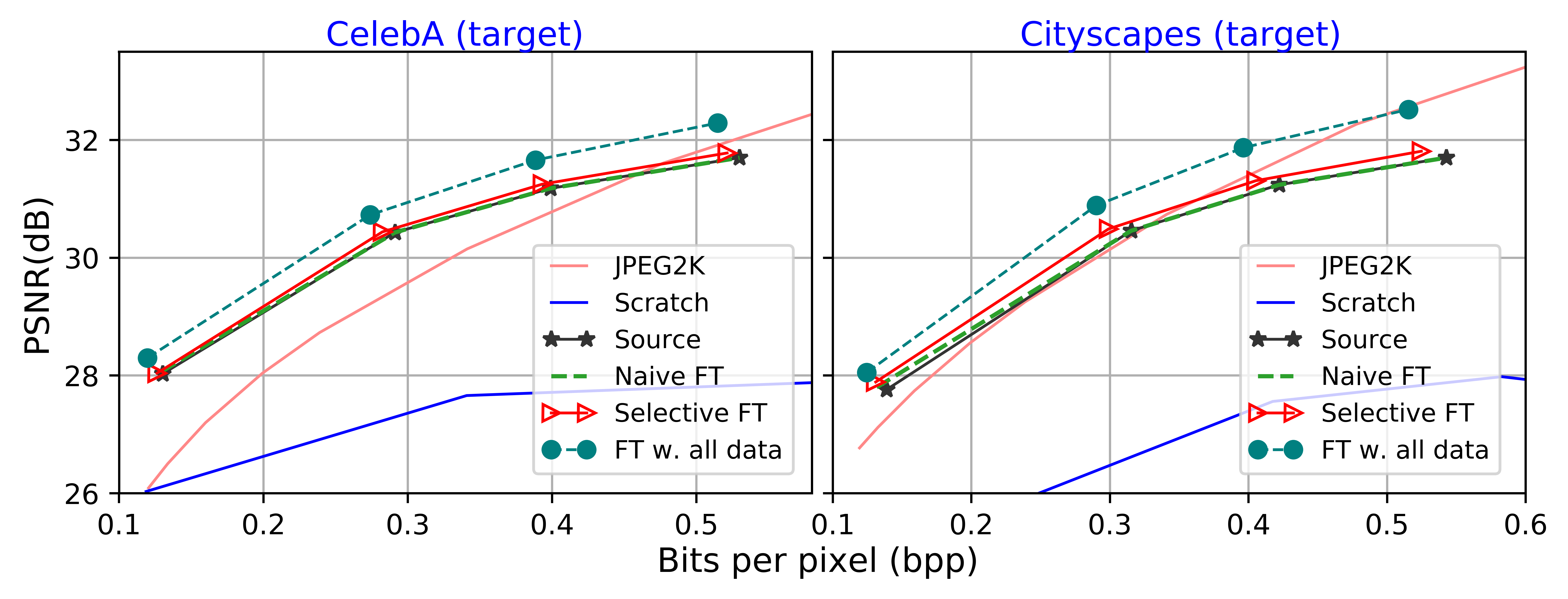}
	\caption{Domain adaptation with 25 images of target domain (CLIC $\rightarrow$ CelebA/Cityscapes).}
	\label{fig:celeba_city_transferability}
\end{figure}

\begin{figure}[t]
\vspace{-0.8em}
    \centering
         \includegraphics[width=\columnwidth]{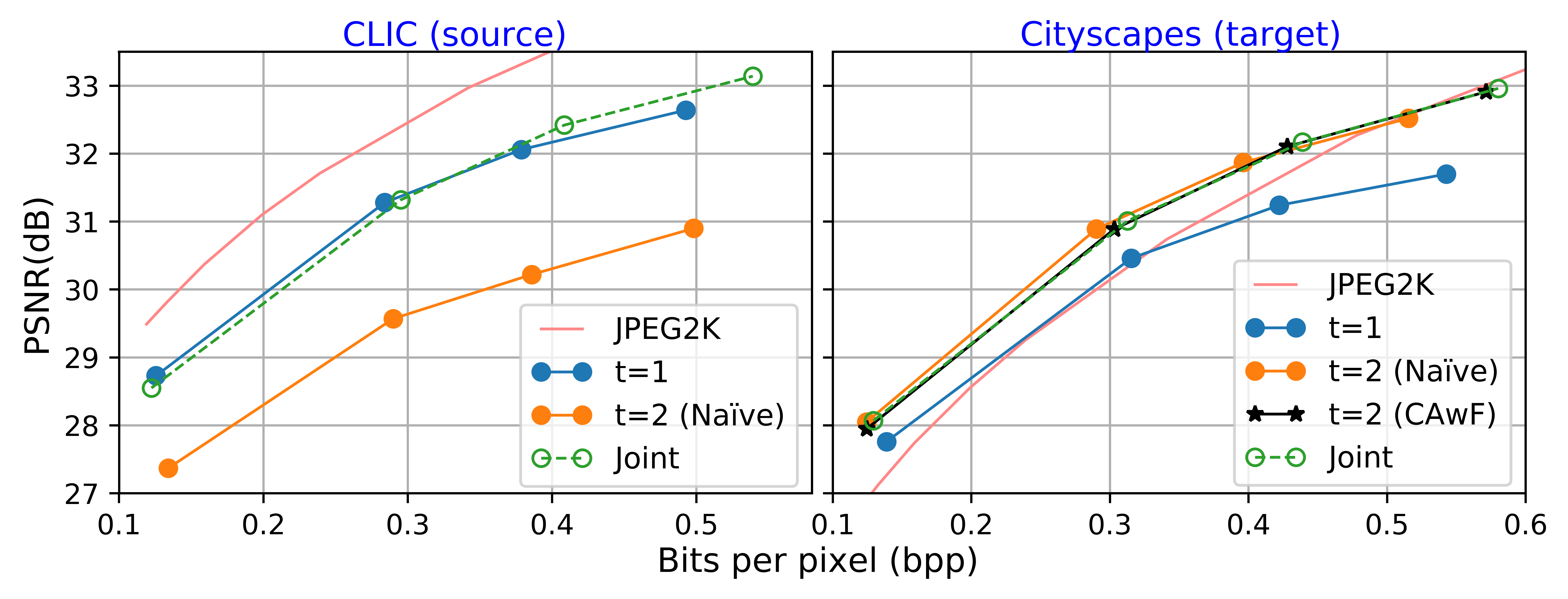}
	\caption{Changes in rate-distortion curves from $t\!=\!1$ to $t\!=\!2$ in both domains for the different experiments (left: source, right: target). Note: on the left figure $t\!=\!2$ (CAwF) corresponds to $t\!=\!1$.
	}
	\label{clic_celeba/city}
	\vspace{-0.8em}
\end{figure}

\begin{table}
\centering
  \caption{Average difference in bitrate due to adaptation to CelebA/Cityscapes with different number of images from target domain (measured as BD-rate (\%) with respect to model obtained by fine tuning source model using all target data).}
  \label{tab:BD-rate_adaptation}
  \resizebox{\columnwidth}{!}{
  \begin{tabular}{c|cc|cc}
    \hline
    &\multicolumn{2}{c|}{CLIC $\rightarrow$ CelebA} &\multicolumn{2}{c}{CLIC $\rightarrow$ Cityscapes} \\
    
    \hline
    Source model &\multicolumn{2}{c|}{19.24} &\multicolumn{2}{c}{23.93}\\
    \hline
    Number of &Naive &Selective &Naive &Selective \\
    target images & fine tuning &fine tuning &fine tuning &fine tuning \\
    \hline
    10 &19.24 &\textbf{16.46} &22.96 &\textbf{17.54}\\
    25 &18.76 &\textbf{14.93} &18.44 &\textbf{15.79}\\
    50 &15.59 &\textbf{13.73} &16.29 &\textbf{15.33}\\
    \hline
    JPEG2K &\multicolumn{2}{c|}{46.96} &\multicolumn{2}{c}{24.55}\\
    \hline
    \end{tabular}
    }
\end{table}

\begin{table}
\centering
  \caption{Average differences in bitrate of various DANICE approaches when forgetting is a concern (measured as BD-rate (\%) with respect to the model trained jointly using source and target data). Lower the value, better the performance.}
  \label{tab:BD-rate_continual}
  \resizebox{\columnwidth}{!}{
  \begin{tabular}{c|cc|c|cc|c}
    \hline    
     & \multicolumn{3}{c|}{CelebA$\rightarrow$Cityscapes} & \multicolumn{3}{c}{Cityscapes$\rightarrow$CelebA} \\
    \hline
    Time & CelebA & Cityscapes & Avg.  & Cityscapes & CelebA & Avg.\\
    \hline
    1     & \textbf{-1.14} & 32.77 & 15.82  & \textbf{-4.75} & 84.32 &39.79 \\
    2 (naive) & 96.25 & \textbf{-4.79} &45.73  & 26.25 & \textbf{-2.45} &11.90 \\
    2 (CAwF) & \textbf{-1.14} & -0.40 & \textbf{-0.77} & \textbf{-4.75} &4.01 & \textbf{-0.37} \\
    \hline
    JPEG2K & 38.26 & 16.34 & 27.3  &16.34  &38.26  &27.3 \\
    \hline 
    \hline
     & \multicolumn{3}{c|}{CLIC$\rightarrow$CelebA} & \multicolumn{3}{c}{CLIC$\rightarrow$Cityscapes}\\    
    \hline
    Time & CLIC & CelebA & Avg. & CLIC & Cityscapes & Avg. \\
    \hline
    1     &\textbf{5.97} &15.59 &10.78 &\textbf{-2.13} &18.90 &8.39\\
    2 (naive) &10.59 &\textbf{-2.05} &4.27 &74.28 &\textbf{-3.18} &35.55\\
    2 (CAwF) &\textbf{5.97} &1.14 &\textbf{3.56} &\textbf{-2.13} &0.27 &-0.93 \\
    \hline
    JPEG2K &-23.57 &40.66 &8.54 &-28.37 &18.38 &\textbf{-4.99} \\
    \hline 
    \end{tabular}
    }
    \vspace{-0.8em}
\end{table}

\begin{figure*}[t]
	\centering
	\small
	\resizebox{0.93\textwidth}{!}{
	\begin{tabularx}{\textwidth}{Y|Y|YY|YY|Y}
	     \multirow{2}{*}{Original} & \multirow{2}{*}{Source} & \multicolumn{2}{c|}{10 images} &  \multicolumn{2}{c|}{50 images} & \multirow{2}{*}{FT with all target data}\\
	     & & Scratch & Selective FT 
 & Scratch & Selective FT &
	\end{tabularx}
 	}
 	\includegraphics[width=0.93\textwidth]{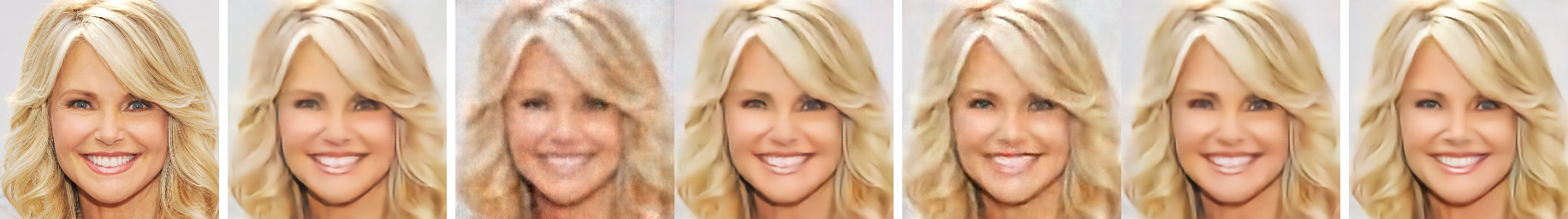}\\
	(a)
	
	\resizebox{0.93\textwidth}{!}{
	\begin{tabularx}{\textwidth}{Y|YY|YYY|YYY|YYY}
	     \multirow{2}{*}{Original} & \multicolumn{2}{c|}{Source} &  \multicolumn{3}{c|}{naive FT (catastrophic forg.)} & \multicolumn{3}{c|}{naive FT (RD forgetting)} & \multicolumn{3}{c}{CAwF}\\
	     & $t=1$ & Error  & $t=2$ & Error & Interf. & $t=2$ &  Error & Interf.& $t=2$ &  Error & Interf.
 	\end{tabularx}	
 	}
	\includegraphics[width=0.93\textwidth]{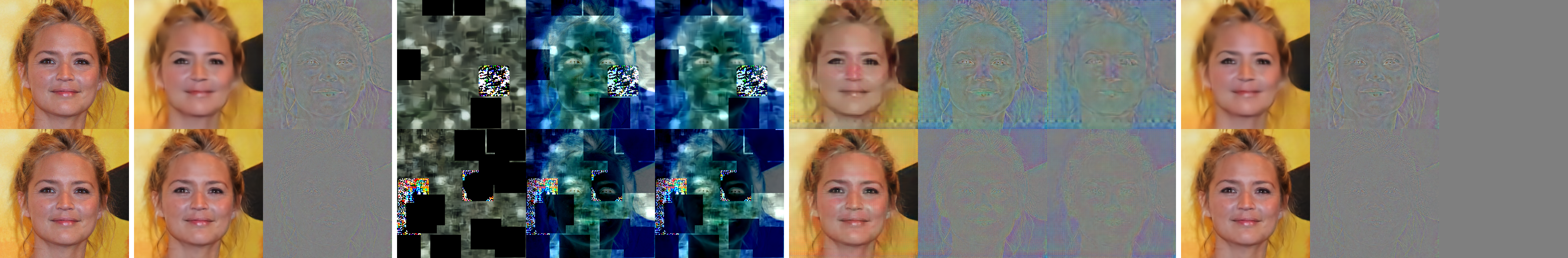} \\
    (b)

\caption{Examples showing that the source model (CLIC) is transferable to CelebA (a), and the benefit of selective fine tuning.
Adaptation from CelebA to Cityscapes (b) shows learning in Cityscapes and forgetting in CelebA. First and second rows in each block row show results for low and high rate codecs, respectively. Error shows the difference image with respect to Original, while Interf. and Change show the difference with respect to image at $t=1$. 
Best viewed in electronic version with zoom.
}
\label{fig:examples}
\end{figure*}

\section{Experiments}

\subsection{Settings}
We evaluate DANICE with few target images (section~\ref{ssec:transferability}) and  when forgetting is a concern (section~\ref{ssec:forgetting}).  We focus our attention on scenarios involving lightweight models and narrow target domains.

\minisection{Domains}
We consider three datasets CLIC, Cityscapes
(resized to $512\times 256$) and CelebA-HQ
(resized to $256\times 256$) for our experiments. CLIC is a wide domain (photos of a variety of topics), while the domains of CelebA and Cityscapes are narrower (faces and street scenes taken from a car, respectively). We experimented with the following pairs of source-target domains: CelebA-Cityscapes, Cityscapes-CelebA, CLIC-CelebA and CLIC-Cityscapes. 

\minisection{Implementation}
Our framework is based on the \textit{factorized-prior} model from~\cite{balle2018variational}\footnote{https://github.com/tensorflow/compression}, where the encoder has four convolutional layers separated by GDN layers, and the decoder  mirrors them. The model is trained for 500k-750k iterations at four RD tradeoffs, $\Lambda=\left\{\lambda_1=0.002,\lambda_2=0.008,\lambda_3=0.016,\lambda_4=0.032 \right\}$. Our framework has 64 shared filters and 16 custom filters per conv. layer. We include JPEG2K\footnote{https://www.openjpeg.org} RD curves as reference.

\minisection{Metrics}
In addition to the usual average rate (in bits-per-pixel, bpp) and average distortion (in average RGB PSNR) curves, we also use the BD-rate metric for comparisons.
    
\subsection{Adaptation with limited data}
\label{ssec:transferability}
We evaluate the benefits of selective fine tuning for DANICE, by comparing it against training from scratch (randomly initialized parameters), using the source model as is, and naive fine tuning. Fig.~\ref{fig:celeba_city_transferability} shows that a source model trained on CLIC achieves competitive RD performance, and performs better than training from scratch using few images. This demonstrates that NIC codecs are transferable and improve RD performance through adaptation with few images (especially using selective fine tuning), which is not possible with a traditional codec such as JPEG2K. 
Table~\ref{tab:BD-rate_adaptation} shows this gain reporting the BD rate of  various approaches using the codec fine tuned using all target data ($\geq 20000$ images) as reference. 
Fig.~\ref{fig:examples} (a) compares reconstructed images using different models, showing that not using a source model (i.e. from scratch) leads to poor quality and artifacts such as noise when the training data is limited.

\subsection{Forgetting and compatibility}\label{ssec:forgetting}

\minisection{Rate-distortion performance}
Fig.~\ref{clic_celeba/city} shows the RD curves on source and target domains. We can observe that after naive fine tuning (\textit{i.e}. $t=2$), the RD performance on the target domain improves (mostly reducing distortion, and slightly reducing rate) while in the source domain often deteriorates significantly, which showcases forgetting. With CAwF, we are able to avoid forgetting in the source domain by design, while being able to attain good performance in the target domain due to the addition of custom parameters. The amount of improvement and decline after adaptation depends on the specific pair of domains, being small for CLIC-CelebA, and large for Cityscapes-CelebA. Table~\ref{tab:BD-rate_continual} reports the BD-rate savings of various approaches with respect to a codec with 80 (64+16) filters trained jointly with both domains, since it often serves as a (soft) upper bound for sequential training. With CAwF, the average of BD-rate of source and target domains is close to zero and much lower than naive fine tuning. 
However, we note that the addition of 16 filters in CAwF can increase the number of parameters up to 55\% (see Table~\ref{tab:parameter_count} in the appendix).

\minisection{Qualitative analysis}
Fig.~\ref{fig:examples} (b) shows several decoded images from the CelebA-Cityscapes experiments. While in all cases images are decodable, we observe that the CelebA bitstreams from $t=1$ turn into a collage of random patches, and encoding at $t=2$ results in a significant loss of RD performance, with noticeable artifacts caused by the interference. In contrast, CAwF is able to prevent interference and keep the original quality. 

\section{Conclusion}
In this paper, we propose a framework to optimize a NIC codec to a target domain of interest, which can be performed directly by the user with a custom dataset. In order to avoid interference that could break the compatibility or harm the coding capabilities in the source domain, target domain knowledge is stored as separate custom parameters. Our work draws from insights from several machine learning areas, chiefly transfer learning, domain adaptation and continual learning, which could provide inspiration for new functionalities in future image and video coding standards. 

\section*{Acknowledgments}
We acknowledge the support from Huawei Kirin Solution and the Spanish Government funding for projects RTI2018-102285-A-I00 and RYC2019-027020-I.

{\small
\bibliographystyle{ieee_fullname}
\bibliography{clic.bib}
}

 \clearpage
	\renewcommand{\appendixpagename}{Supplementary material}
	\begin{appendices}

\section{Details of NIC codecs}
Encoding consists of a learnable nonlinear mapping $\mathbf{z}=\tilde{f}\left(\mathbf{x};\theta\right)$ where the feature encoder is parametrized by $\theta$ that maps an input image $\mathbf{x}\in \mathbb{R}^{N}$ to a latent representation $\mathbf{z}\in\mathbb{R}^{M}$.
However, the latent representation is still redundant and in continuous space, which is not amenable to transmission through digital communication channels. Thus, the latent representation is quantized as
$\mathbf{q}=Q\left(\mathbf{z}\right)$, where $\mathbf{q}\in \mathbb{Z}^M$ is a discrete-valued symbol vector (in this paper we use quantization to the nearest integer, i.e. $\mathbf{q}=\lfloor\mathbf{z}\rceil$). Finally, the symbol vector $\mathbf{q}$ is binarized and serialized into the bitstream $\mathbf{b}$, using a (lossless) entropy encoder that exploits its statistical redundancy.
Entropy coding is reversed during the process of decoding to obtain $\mathbf{q}$, which is then processed by the feature decoder in order to obtain the reconstructed image. This is be summarized by $\mathbf{\hat{x}}=\tilde{g}\left(\mathbf{q};\phi\right)$, where $\mathbf{\hat{x}}\in \mathbb{R}^{N}$ is the reconstructed image, and the feature decoder, $\tilde{g}$ is parametrized by $\phi$. 
In particular, we follow the framework of Balle \textit{et al.}~\cite{balle2016end,balle2018variational}, which combines convolutional layers, generalized divisive normalization (GDN)~\cite{balle2016density} and inverse GDN layers, scalar quantization and arithmetic coding. 

During training, quantization is replaced by a differentiable proxy to allow end-to-end training via backpropagation~\cite{balle2016end}. In this paper, additive uniform noise $\mathbf{\tilde{z}}=\mathbf{z}+\Delta \mathbf{z}$, with $\Delta \mathbf{z}\sim \mathcal{U}\left(-\frac{1}{2},\frac{1}{2}\right)$ is used as the proxy. Rate is estimated as the entropy of the quantized symbol vector $R\left(\mathbf{b}\right)\approx H\left[P_\mathbf{q}\right]\approx H\left[p_\mathbf{\tilde{z}}\left(\mathbf{\tilde{z}};\nu\right)\right]$, where $\nu$ are the parameters of the entropy model. 
The entropy model is known by both encoder and decoder. After training, the full model is thus determined by parameters $\psi=\left(\theta,\phi,\nu\right)$, with the full (including arithmetic coding) encoder $f\left(\mathbf{x};\theta,\nu\right)$ and decoder $g\left(\mathbf{b};\phi,\nu\right)$.

The parameters $\psi$ are learned by minimizing a combination of rate and distortion over a training set $\mathcal{X}^\text{tr}$ sampled from the domain of interest $\mathcal{X}$ 
\begin{gather}
	J\left(\mathcal{X}^\text{tr},\psi;\lambda\right)=R\left(\mathcal{X}^\text{tr},\psi\right)+\lambda D\left(\mathcal{X}^{tr},\psi\right) \label{eq:RD_loss_appx},
\end{gather}
where $\lambda$ is the (fixed by design) tradeoff between rate and distortion. 

In this paper, distortion is measured as the average reconstruction mean square error (MSE)
\begin{equation}
	D\left(\mathcal{X}^\text{tr},\psi\right)=\mathbb{E}_{\mathbf{x}\in\mathcal{X}^\text{tr}}\left\|\mathbf{\hat{x}}-\mathbf{x}\right\|^2
\end{equation}
and rate estimated as the average entropy
\begin{equation}
	R\left(\mathcal{X}^\text{tr},\psi\right)=\mathbb{E}_{\mathbf{x}\in\mathcal{X}^\text{tr}} H\left[p_\mathbf{\tilde{z}}\left(\mathbf{\tilde{z}; \nu}\right)\right]. 
\end{equation}

\section{Related work}
\minisection{Image compression} 
Widely used image and video coding standards are based on the successful combination of prediction of pixels in small blocks, linear transforms (typically DCT), quantization and entropy coding~\cite{wallace1992jpeg}. Most advanced standards include carefully designed prediction and coding tools that greatly improve coding efficiency and quality~\cite{wiegand2003overview,sullivan2012overview,vvc2020}. 
Due to the pervasiveness of strictly defined decoder implementations in consumer devices, backward compatibility with legacy standards has always been a desirable characteristic in applications using image and video coding. A consequence of such approach is very limited adaptability of underlying codecs, leading to suboptimal use of compression technology in many application domains.

In contrast, NIC uses highly flexible parametric architectures whose parameters are learned from data. A typical architecture consists of a deep autoencoder followed by quantization and entropy coding (using differentiable proxies during optimization)~\cite{theis2017lossy,balle2016end,balle2018variational,toderici2017full}. Recent methods include hyperpriors~\cite{balle2018variational} and autoregressive probability models~\cite{toderici2017full} to improve performance. However, the application of adapting learned image codecs to new domains while keeping the performance and compatibility with the original domain has not been considered.

\minisection{Transfer learning and domain adaptation}
Deep neural networks (pre)trained on large datasets, can be reused and adapted (e.g. fine tuned) to specific target tasks, even with limited data (i.e.transfer learning)~\cite{azizpour2015factors}. Domain adaptation~\cite{wang2018deep} is the specific case when the task remains the same across domains (e.g. source and target tasks have the same categories). 
Here, we address domain adaptation in the context of image compression.

\minisection{Continual learning}
Continual learning~\cite{parisi2019continual} addresses challenging scenarios where the model has to learn from continually arriving non-i.i.d. data from new unseen categories, domains or tasks. The most characteristic problem in this setting is the (catastrophic) forgetting~\cite{mccloskey1989catastrophic} of previously learned skills, due to the interference between previous domains or tasks. It is often addressed using specific regularization, (pseudo)rehearsal of previous data or using task-specific parameters. Here, we study and address interference and forgetting in NIC.

\begin{table}
    \centering
    \caption{Comparison of the number of parameters between base model and \textit{CAwF}}
    \label{tab:parameter_count}
    \resizebox{\columnwidth}{!}{
    \begin{tabular}{|c|c|ccc|c|}
        \hline
        & \multirow{2}{*}{Filters} & \multirow{2}{*}{Encoder} & Entropy & \multirow{2}{*}{Decoder} & Total  \\ & & & model & & parameters  \\
        \hline 
        Base & 64 & 324736 & 6208 & 324675 & 655619  \\
        \hline 
        \textit{CAwF} & 80 & 505760 & 7760 & 505683 & 1019203 \\
        \hline 
    \end{tabular}
    }
    
\end{table}

\end{appendices}

\end{document}